\documentclass[letterpaper]{article}
\usepackage[preprint]{aaai2027}
\usepackage[hyphens]{url}
\usepackage{graphicx}
\graphicspath{{figures/}{Figures/}}
\usepackage{natbib}
\usepackage{caption}
\usepackage{amsmath}
\usepackage{amssymb}
\usepackage{booktabs}

\usepackage{multirow}

\newcommand{\method}{ET-Prune}
\newcommand{\mllm}{MLLM}
\newcommand{\mllms}{MLLMs}

\title{ET-Prune: Evidence-Aware Dynamic Budgeting for Visual Token Pruning in Text-Rich MLLMs}
\author{
    Zizhong Ding\textsuperscript{\rm 1}\equalcontrib
    Junxian Li\textsuperscript{\rm 1}\equalcontrib\thanks{Project leader.},
    Kai Liu\textsuperscript{\rm 1}, \\
    Shaoqiu Zhang\textsuperscript{\rm 1}, 
    Xiao Xiao\textsuperscript{\rm 2},
    Linghe Kong\textsuperscript{\rm 1}, 
    Yulun Zhang\textsuperscript{\rm 1}\corresponding
}
\affiliations{
    \textsuperscript{\rm 1}Shanghai Jiao Tong University,
    \textsuperscript{\rm 2}Xidian University \\
    \small{\texttt{yulun100@gmail.com}}

}

\begin{document}
\maketitle

\begin{abstract}
Visual token pruning reduces the inference cost of multimodal large language
models, but a fixed token ratio is poorly matched to text-rich inputs. In
OCR-centric tasks, decisive evidence can be a small number, label, or field
whose relevance is specified by the question; indiscriminate pruning can erase
that evidence while retaining visually salient but irrelevant regions. We
present \method, a training-free framework that casts pruning as \emph{evidence
allocation}. It derives question-conditioned evidence from a decoder-side
partial query--key block, safeguards text-like spatial regions, and converts
evidence uncertainty and density into a sample-specific token floor. Three
progressive middle-layer events then move the sequence toward this budget,
retaining more tokens for diffuse or text-dense evidence and pruning concentrated
evidence more aggressively. At the observed point estimates from one
deterministic pass per configuration, \method{} leads or ties among pruned
methods in all six backbone--benchmark comparisons at roughly half tokens. On
OCRBench-v2, it leads the strongest pruned baselines by 1.80 and 0.68
percentage points on Qwen3-VL-8B and InternVL3.5-8B, respectively, while
retaining about half of the visual tokens; on MMBench v1.1, it reaches 0.8467
circular exact-matching accuracy versus 0.8437 for Vanilla at 54.45\% average
visual-token retention. These results show a favorable observed quality--cost
trade-off for evidence-aware dynamic budgeting in text-rich multimodal
inference. Code is at https://github.com/Labyrinth0419/ET-Prune.
\end{abstract}

\section{Introduction}

\mllms{} have become effective tools for OCR-centric tasks, including scene-text
question answering, document understanding, chart reasoning, and screen
reasoning~\cite{singh2019textvqa,mathew2021docvqa,masry2022chartqa,
ye2023ureader,hu2024docowl15,li2024monkey,li2025chemvlm,kang2026hssbench,gao2026laobench}. However, their ability to resolve
small and densely arranged text relies on high-resolution visual inputs, which
generate long visual-token sequences and increase decoder computation,
KV-cache memory, and prefill latency, especially on resource-constrained edge
devices. These costs have consequently motivated extensive efforts to
accelerate inference and reduce computational overhead.


Visual token pruning directly reduces this overhead by removing redundant
visual tokens. Existing methods typically rank tokens using generic attention,
feature magnitude, or redundancy criteria
~\cite{chen2024fastv,xing2025pyramiddrop,zhang2025sparsevlm,alvar2025divprune,kong2025token,li2026g},
and often impose a fixed keep ratio. These policies work well when semantic
evidence is concentrated in normal scenarios, but they may fail under text-rich
inputs. We observe three critical challenges on these tasks. Firstly, the text region directly related to the answer may be \textbf{small and visually inconspicuous}, making it
easy for generic saliency to discard. Secondly, \textbf{token relevance changes largely} for different instructions, like identifying a color and transcribing a receipt total needs very different visual regions. Thirdly, the amount of \textbf{evidence that must
be preserved varies} across inputs. A global keep ratio can over-allocate
tokens to easy examples while under-protecting difficult ones. These
observations motivate question-conditioned scoring and sample-specific token
budgets. A posterior analysis of the full-token model further shows that
answer-conditioned visual attention can be diffuse or concentrated. 

\begin{figure*}[t]
    \centering
    \includegraphics[width=\textwidth]{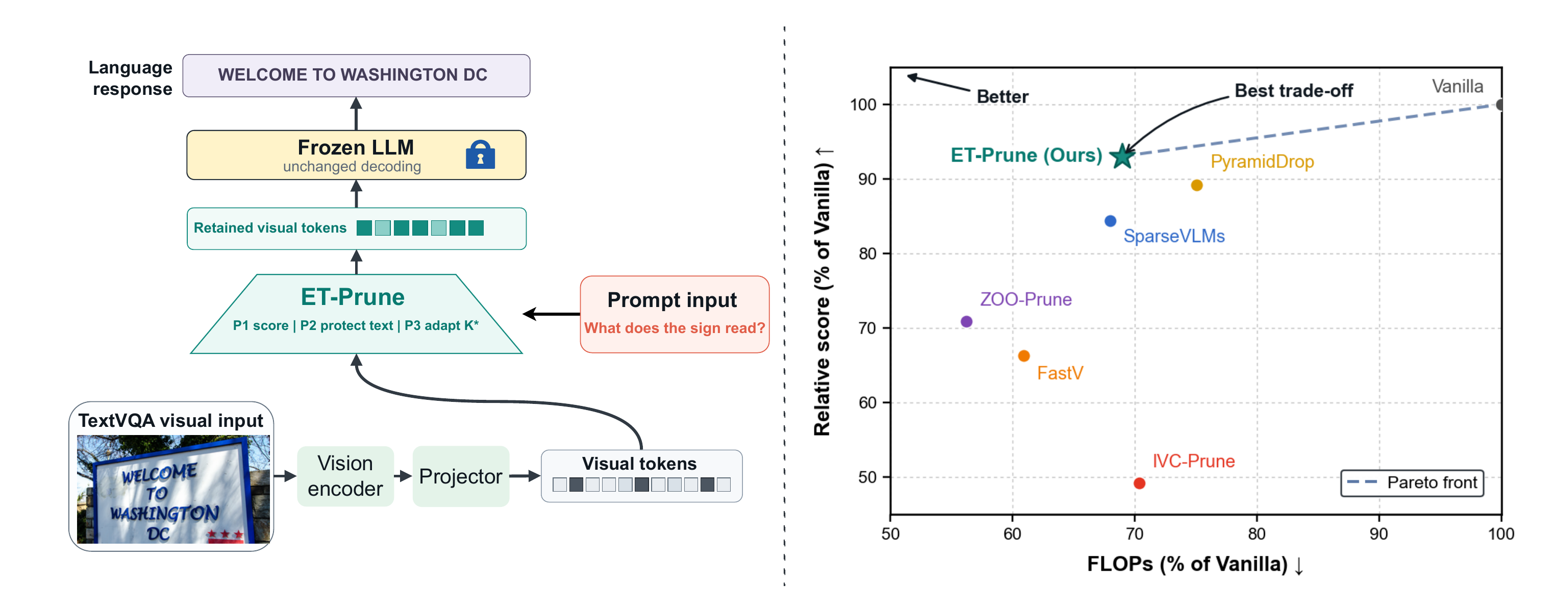}
    \caption{Overview and quality--cost trade-off of \method{}. Left:
    question-conditioned evidence scoring, text-region safeguarding, and
    sample-specific budgeting before frozen decoding. Right: on Qwen3-VL-8B
    and OCRBench-v2, \method{} retains 93.0\% of the Vanilla score at 69.0\%
    FLOPs.}
    \label{fig:teaser}
\end{figure*}

\begin{figure*}[t]
    \centering
    \includegraphics[width=\textwidth]{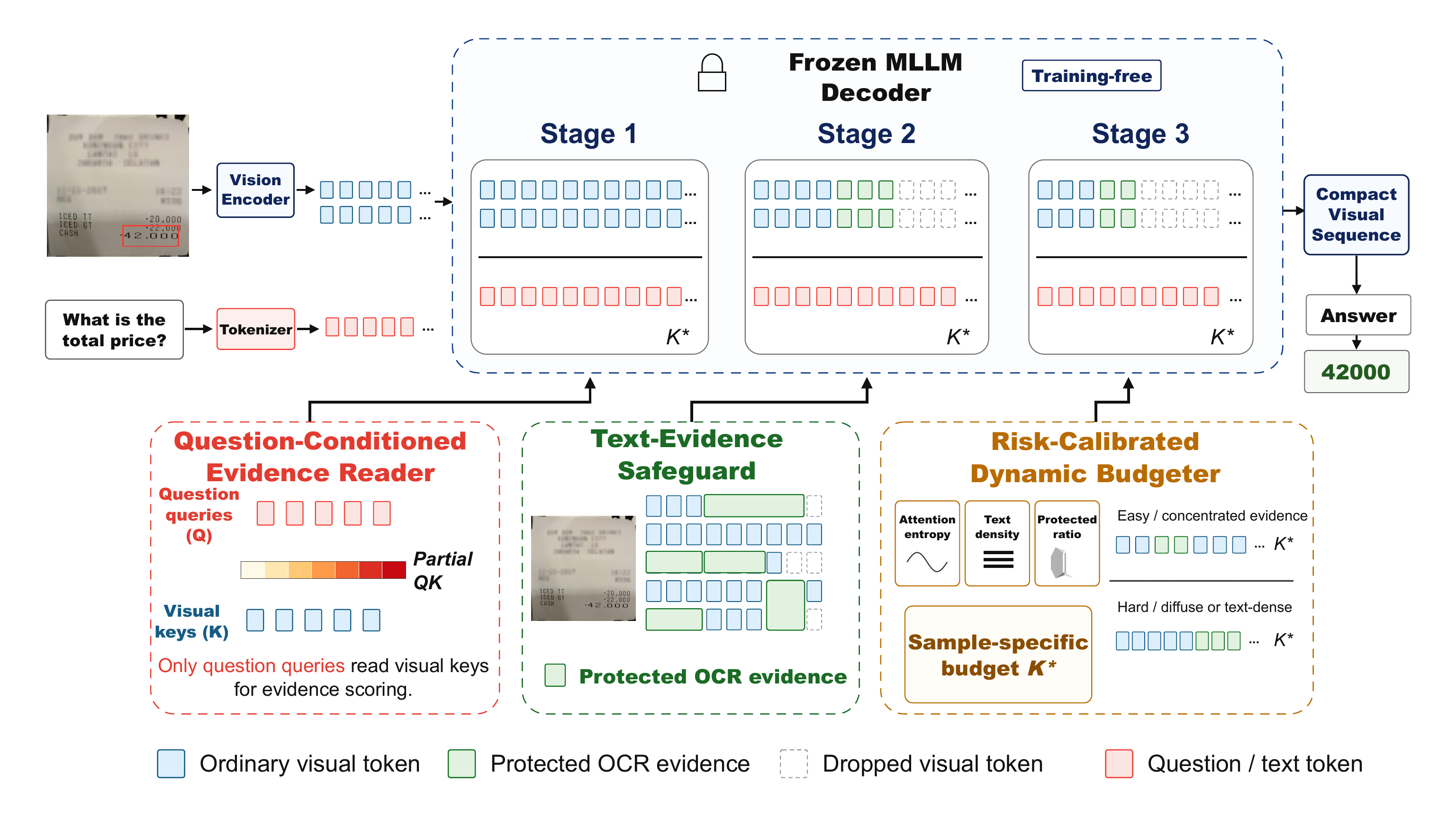}
    \caption{Overview of \method{}. A partial-QK reader scores
    question-conditioned evidence, a spatial safeguard protects text-like
    regions, and a risk-calibrated budgeter sets a sample-specific token floor.
    Three middle-layer stages progressively compact the visual sequence.}
    \label{fig:overview}
\end{figure*}

\begin{figure}[t]
    \centering
    \includegraphics[width=\columnwidth]{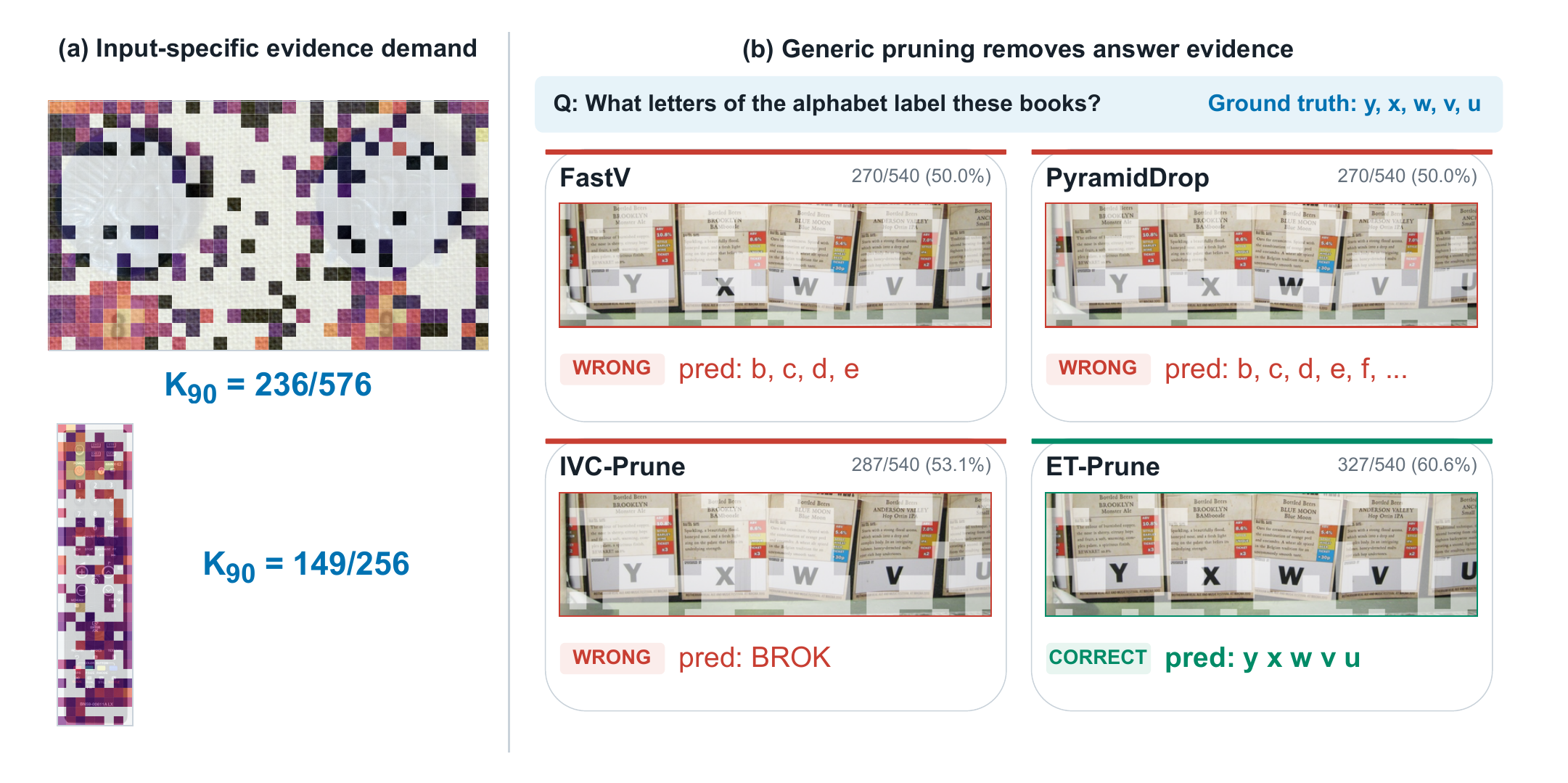}
    \caption{Motivating examples. (a) Full-token posterior
    answer-conditioned visual attention. (b) FastV, PyramidDrop, IVC-Prune, and \method{}
    retain 50\%, 50\%, 53\%, and 61\% of the tokens, respectively; the first
    three obscure answer-bearing labels in this example, while \method{}
    preserves them.}
    \label{fig:pruning-motivation}
\end{figure}

\begin{figure}[t]
    \centering
    \includegraphics[width=0.86\columnwidth]{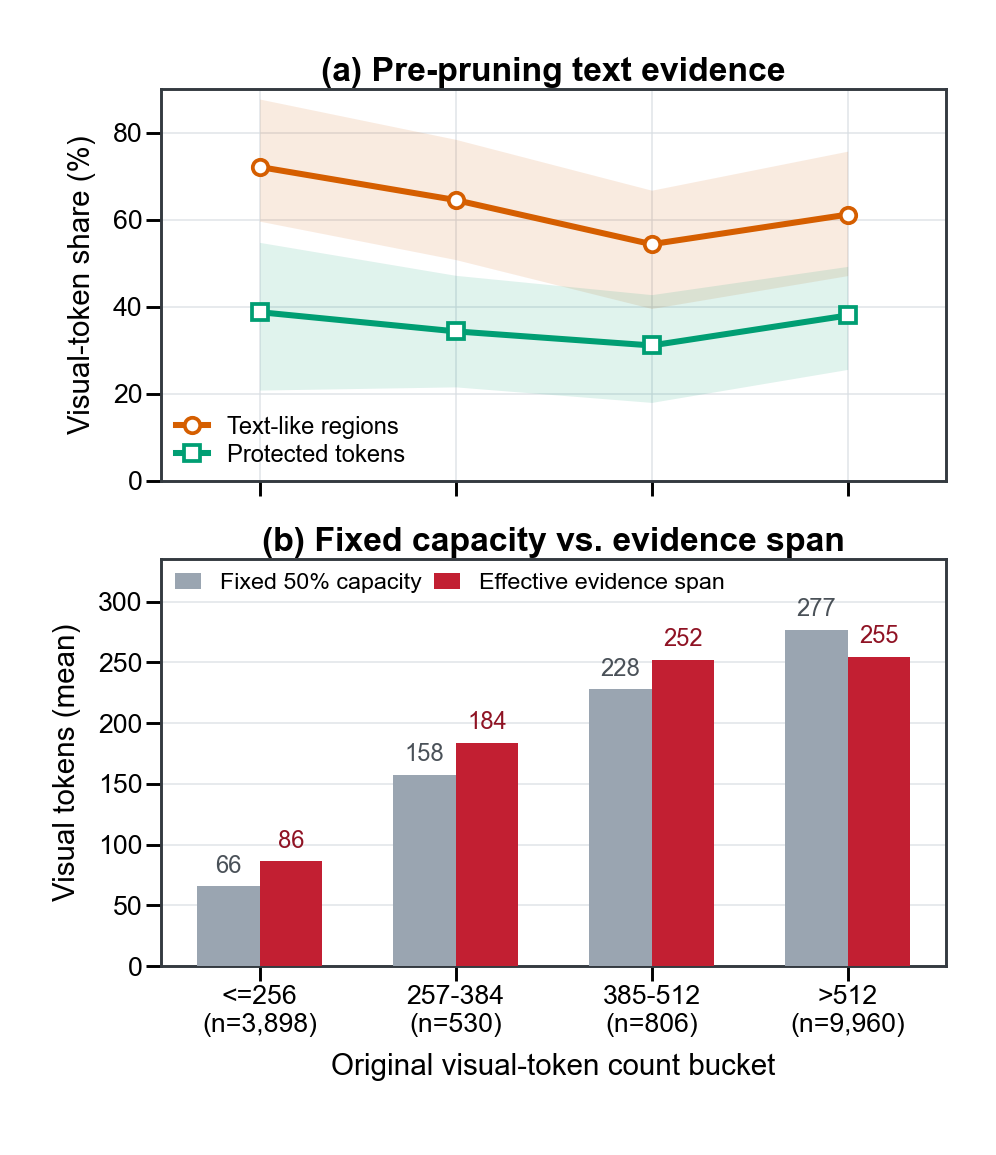}
    \caption{Evidence heterogeneity across 15,194 Qwen3-VL inputs
    grouped by visual-token count. (a) Text-like and protected-token fractions
    vary with input scale; bands show interquartile ranges. (b) The fixed 50\%
    budget diverges from the question-conditioned effective evidence span
    across scales.}
    \label{fig:motivation}
\end{figure}

To address these challenges, we formulate OCR-oriented token pruning as a
question-aware evidence-allocation problem. The goal is to preserve the
evidence needed to answer the current question within a limited token budget,
prioritizing answer relevance during token selection. This formulation leads to
three design principles: (i) token importance should incorporate
question-to-visual affinity; (ii) small, text-like regions should receive
explicit protection because their evidence is fragile under patch-level
removal; and (iii) the retained token budget should be determined per example,
with more capacity allocated when the evidence is uncertain or text-dense.
Together, these principles turn visual token pruning into a question-aware
allocation problem and motivate the training-free framework introduced next.

We instantiate these principles in \textbf{\method{}}, a training-free framework for
text-rich \mllms{}. Its
question-conditioned evidence reader extracts
question-to-visual affinity from a decoder-side partial-QK block; its
text-evidence safeguard maps lightweight spatial cues to text-like components
and protects their token support; and its risk-calibrated budgeter converts
evidence uncertainty, text density, and protected-region prevalence into a
sample-specific token floor. Progressive middle-layer compaction then moves
each visual sequence toward this floor while keeping the model weights and
decoding procedure frozen. Across two core backbones and three text-rich
benchmarks, \method{} leads or ties the pruned baselines in all six
backbone--benchmark comparisons at roughly half of the visual tokens. On the
representative OCRBench-v2 operating point in Figure~\ref{fig:teaser}, it
retains 93.0\% of the Vanilla score at 69.0\% of Vanilla FLOPs, illustrating
the resulting trade-off between accuracy and efficiency.
This paper makes the following contributions:
\begin{itemize}
    \item We conduct an in-depth study of evidence heterogeneity in OCR-centric
    visual token pruning, examining how answer relevance and required token
    capacity vary across questions and input scales. This analysis highlights
    the limitations of a fixed global keep ratio and motivates
    question-aware evidence allocation.

    \item We propose \method{}, a training-free framework that realizes
    evidence allocation through a decoder-side partial-QK reader, a spatial
    safeguard for fragile text-like regions, risk-calibrated sample-specific
    budgeting, and progressive middle-layer compaction.

    \item We evaluate \method{} on two core \mllm{} backbones and three
    text-rich benchmarks under matched pruning budgets. It leads or ties all
    pruned baselines across the six core backbone--benchmark comparisons while
    retaining roughly half of the visual tokens. Additional analyses examine
    pruning depth, model scaling, inference efficiency, and cross-benchmark
    transfer.
\end{itemize}

\section{Related Work}


\paragraph{Text-rich multimodal understanding.}
Text-rich multimodal understanding requires recognizing fine-grained text and
grounding it in visual context. TextVQA, DocVQA, and ChartQA evaluate scene
text, documents, and charts, while OCRBench-v2 adds bilingual localization,
parsing, knowledge, and reasoning~\cite{singh2019textvqa,mathew2021docvqa,
masry2022chartqa,fu2025ocrbenchv2}. OCR-oriented \mllms{} such as UReader,
mPLUG-DocOwl, and Monkey improve these capabilities through high-resolution
encoding and specialized training~\cite{ye2023ureader,hu2024docowl15,
li2024monkey}. These design choices lengthen visual sequences and increase
inference cost, creating a need for efficient processing of text-rich inputs.

\paragraph{Adaptive computation.}
Adaptive computation allocates capacity according to input difficulty.
BlockDrop and DynamicViT learn conditional execution for network blocks and
vision tokens, while CALM and Mixture-of-Depths adapt language-model
computation~\cite{wu2018blockdrop,rao2021dynamicvit,schuster2022calm,
raposo2024mixturedepths}. These methods establish input-dependent allocation as
a general efficiency principle. In frozen text-rich \mllms{}, lightweight
evidence signals can guide adaptive visual computation without retraining the
backbone.

\paragraph{Visual token compression and pruning.}
Visual-token reduction lowers decoder cost through token merging or pruning.
We focus on five training-free pruning baselines that differ in their selection
signals and schedules. FastV uses early text-to-vision attention for one-shot
removal, and SparseVLMs performs text-relevance-guided multi-layer
sparsification~\cite{chen2024fastv,zhang2025sparsevlm}. PyramidDrop
progressively prunes tokens with depth~\cite{xing2025pyramiddrop}. IVC-Prune
uses implicit visual coordinates~\cite{sun2026ivcprune}, and ZOO-Prune
estimates output sensitivity through zeroth-order
perturbations~\cite{kim2025zooprune}. Together, they span attention, relevance,
depth schedule, coordinate, and sensitivity signals. We study question-aware
token selection together with per-instance capacity allocation for text-rich
inputs.

\section{Preliminaries}

\subsection{Visual Token Pruning in MLLMs}

Given an image $I$ and an input prompt $x$, a vision encoder and connector
produce $N$ visual tokens $V=\{v_i\}_{i=1}^{N}$. These tokens are inserted into
the language-model sequence together with text tokens $X$. At a decoder layer
$\ell$, a pruning method selects a subset $\widehat{V}^{(\ell)} \subseteq
V^{(\ell)}$ and propagates the compacted sequence through the remaining layers.
Let $K$ be the final number of retained visual tokens. The usual fixed-budget
formulation sets $K=\rho N$ for a global keep ratio $\rho$. In contrast, our goal
is to choose $K(I,x)$ jointly with the retained subset so that the average visual
budget is controlled without imposing the same compression severity on every
input:
\begin{equation}
    \max \; \mathbb{E}\big[\mathcal{A}(I,x;\widehat{V})\big]
    \quad \text{s.t.} \quad
    \mathbb{E}\big[|\widehat{V}|\big] \leq \bar{K},
    \label{eq:budget-objective}
\end{equation}
where $\mathcal{A}$ denotes task quality and $\bar{K}$ is a desired average
visual-token budget.

\subsection{Motivation of Method}

Figure~\ref{fig:pruning-motivation} illustrates this heterogeneity at the
example level. (We define
$K_{90}$ as the smallest set of top-ranked visual tokens whose cumulative
posterior visual-attention mass reaches 90\%.) It contrasts a large input with localized evidence and a smaller
input with diffuse evidence, and shows how generic pruning can obscure small
answer-related labels.

Fixed-ratio pruning assumes that the same fraction of visual tokens provides a
commensurate evidence budget for every input. This assumption is especially
fragile for text-rich images: a short visual sequence may be dominated by small
glyphs, while a larger sequence may contain substantial background redundancy.
The question further determines whether a sparse field must survive or whether
the image can be compressed aggressively.

A dataset-level analysis makes this mismatch concrete on 15,194 Qwen3-VL inputs
from TextVQA, OCRBench-v2, and OCRBench-mini. All quantities are
available before answer correctness is known. We summarize question-to-visual
affinity by its entropy-equivalent support size, which we call the
\emph{effective evidence span}. In the smallest bucket, a fixed 50\% budget
leaves 66 tokens on average while the evidence span is 86 tokens; in the
largest bucket, it leaves 277 tokens for a 255-token span. A global ratio is
therefore neither uniformly conservative nor uniformly aggressive.

The resulting problem has two coupled directions: identify the visual evidence
relevant to the question, then allocate sufficient capacity to preserve it.
This motivates evidence-aware dynamic budgeting rather than a single global
keep ratio.

\subsection{Why Text-Rich Inputs Need Adaptive Budgets}

Figure~\ref{fig:motivation} further summarizes this scale-dependent mismatch.
The evidence relevant to an OCR-style question is often both localized and
conditional. Let $Q \subseteq X$ denote the question-token span and let $R$ be
the answer-bearing visual region. The desired pruning policy must preserve $R$
when $Q$ requests an exact transcription, a number, or a spatially specified
field, but may safely discard many visual tokens from the same image for a
coarse question. Two signals quantify this uncertainty in our framework: the
concentration of
question-to-visual affinity and the prevalence of text-like visual regions. A
diffuse affinity distribution indicates that several regions may support the
answer; a high text-like density indicates that pruning individual patches is
more likely to fragment an answer-bearing span. These observations motivate the
evidence reader and risk-calibrated budget described next.
\section{Methodology}

\method{} is a training-free, middle-layer pruning framework (Figure~\ref{fig:overview}).
It comprises three cooperating components: a question-conditioned evidence
reader, a text-evidence safeguard, and a risk-calibrated dynamic budgeter.
All three components operate on the current input and frozen \mllm{}
representations, so the model parameters and decoding procedure remain fixed.

\subsection{Framework Overview}

Let $\mathcal I_0=\{1,\ldots,N\}$ denote the initial visual-token indices and
let $\mathcal I_j\subseteq\mathcal I_{j-1}$ be the active indices after pruning
event $j$. Before the language-model prefill, \method{} computes a text-like
spatial prior on the image and maps it to the original visual-token grid. At
the first selected middle layer, the evidence reader estimates the
question-conditioned visual distribution and the budgeter fixes a sample-specific
endpoint $K^*$. At each subsequent event, the reader scores the currently
active visual tokens, the safeguard marks protected support, and the selector
chooses a target-sized nested set. The runtime then gathers hidden states,
attention masks, position information, and cache metadata for the compacted
sequence. The later decoder layers and autoregressive steps operate on this
shorter visual prefix.

\subsection{Question-Conditioned Evidence Reader}

Generic visual saliency cannot distinguish a prominent but irrelevant object
from a small answer-bearing word. We instead use the question as an evidence
query. The runtime identifies the question span $Q$ among the non-visual input
tokens and, at a pruning layer, reads its affinity to the visual keys. Let
$\mathcal{S}$ denote all keys visible to the question queries after the decoder
mask and $\mathcal{V}\subseteq\mathcal{S}$ the current visual keys. For attention
head $h$, we materialize only the question-query block
\begin{equation}
    \mathbf{A}_{Q\rightarrow V}^{(\ell,h)} =
    \left[\operatorname{softmax}_{\mathcal{S}}\left(
      \frac{\mathbf{Q}_{Q,h}\mathbf{K}_{\mathcal{S},h}^{\top}}
      {\sqrt{d_h}}+\mathbf{M}_{Q,\mathcal{S}}
    \right)\right]_{:,\mathcal{V}},
    \label{eq:partial-qk}
\end{equation}
where the softmax normalizes each question row over all mask-permitted keys and
only then slices the visual columns. This avoids materializing a full decoder
attention matrix. Averaging over question tokens and heads yields a visual
evidence score $a_i$ for each token $v_i$:
\begin{equation}
    a_i^{(\ell)}
    = \frac{1}{|Q||\mathcal H|}
      \sum_{q\in Q}\sum_{h\in\mathcal H}
      A_{q,i}^{(\ell,h)},
    \qquad
    \pi_i^{(\ell)}
    = \frac{a_i^{(\ell)}}
      {\sum_{r\in\mathcal V_\ell}a_r^{(\ell)}}.
    \label{eq:evidence-score}
\end{equation}
Here $\mathcal H$ is the set of attention heads and $\mathcal V_\ell$ is the
active visual-token set at layer $\ell$. The first average combines the full
question span and all heads. Since Eq.~\ref{eq:partial-qk} normalizes over all
mask-permitted keys before visual columns are selected, the sliced entries do
not themselves sum to one; the second normalization produces the visual
distribution $\pi^{(\ell)}$ used for entropy and ranking. The score is used for
selection and budget calibration, while the main forward pass remains on the
model's efficient attention path.

\subsection{Text-Evidence Safeguard}

Question affinity alone can be brittle when evidence consists of a few small
glyphs. \method{} computes a lightweight text-like spatial prior before the
language-model prefill. Local contrast and edge responses produce a pixel-level
mask $\mathcal T$ of accepted text-like components, which is projected onto the
visual-token grid. For token region $G_i$, we define
\begin{equation}
    p_i = \frac{|G_i\cap\mathcal T|}{|G_i|},
    \qquad
    m_i = \mathbf 1[|G_i\cap\mathcal T|>0].
    \label{eq:text-prior}
\end{equation}
Here $p_i\in[0,1]$ measures text-like coverage and $m_i$ marks the token
support of a connected component. The safeguard is a training-free structural
prior for sparse, stroke-like regions and performs no text transcription.
Component support is mapped back to the original token indices, so neighboring
patches belonging to the same text line can be protected together at later
pruning events.

At event $j$, let $\mathcal P_j=\{i\in\mathcal I_{j-1}:m_i=1\}$ be the
currently active protected set and let $a_i^{(\ell_j)}$ be the reader score.
The selected set is
\begin{equation}
\mathcal I_j =
\begin{cases}
\mathcal P_j\cup
\operatorname{Top}_{K_j-|\mathcal P_j|}
 (\mathcal I_{j-1}\setminus\mathcal P_j; a^{(\ell_j)}),
 & |\mathcal P_j|\le K_j,\\[2pt]
\operatorname{Top}_{K_j}(\mathcal P_j; a^{(\ell_j)}),
 & |\mathcal P_j|>K_j.
\end{cases}
\label{eq:protected-selection}
\end{equation}
Here $\operatorname{Top}_k(\mathcal A;a)$ returns the $k$ indices in
$\mathcal A$ with the largest scores under $a$. Thus protected support receives
priority, and evidence scores resolve the remaining slots or the rare
protected-set overflow case.

\subsection{Risk-Calibrated Dynamic Budgeting}

\begin{figure}[t]
    \centering
    \includegraphics[width=0.85\columnwidth]{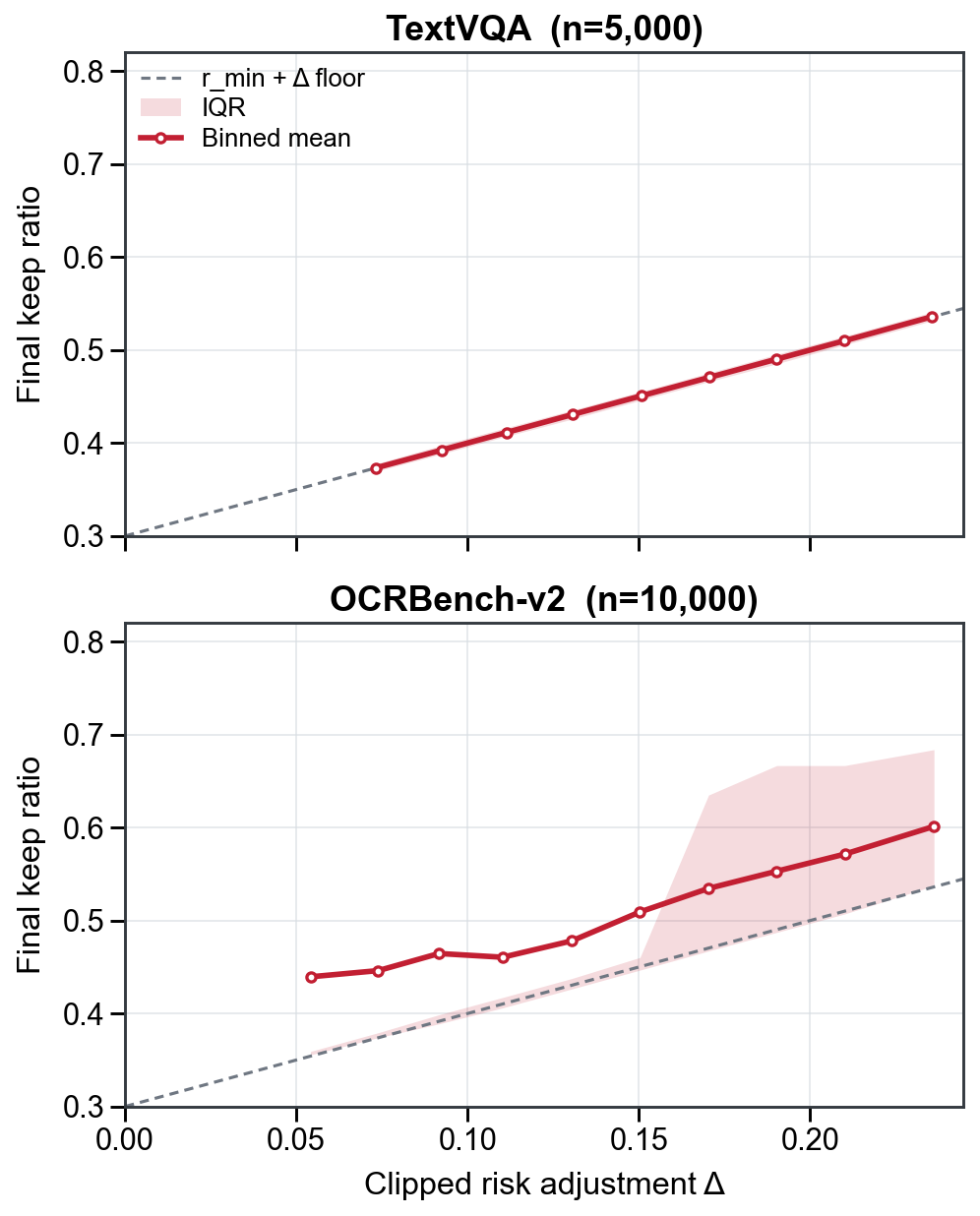}
    \caption{Dynamic budget response on Qwen3-VL. Red: binned mean; shading:
    interquartile range; gray dashed line: risk-adjusted floor
    $r_0+\Delta$. Larger risk adjustments yield higher token retention on
    TextVQA ($n=5{,}000$) and OCRBench-v2 ($n=10{,}000$).}
    \label{fig:budget-risk}
\end{figure}

Token pruning requires decisions about both which tokens to retain and how much
capacity to allocate. We compute the following signals from the initial visual
grid and the question-conditioned distribution $\pi$:
\begin{equation}
\begin{aligned}
    H(a) &= -\frac{1}{\log N}\sum_{i=1}^{N}\pi_i\log\pi_i,\\
    D_p &= \frac{1}{N}\sum_{i=1}^{N}p_i,
    \qquad
    R_p = \frac{1}{N}\sum_{i=1}^{N}m_i.
\end{aligned}
\label{eq:risk-signals}
\end{equation}
The normalized entropy $H(a)\in[0,1]$ is high when question evidence is
distributed across many visual tokens. $D_p$ measures text-like coverage and
$R_p$ measures the fraction of tokens supporting protected components. We
combine them into an evidence-risk adjustment:
\begin{equation}
    \Delta(I,x) = \min\left\{\Delta_{\max},
    w_h H(a) + w_d D_p + w_r R_p\right\}.
    \label{eq:risk}
\end{equation}
Starting from a calibrated base ratio $r_0$, the effective ratio and integer
sample budget are
\begin{equation}
\begin{aligned}
    r_{\mathrm{eff}}(I,x)
      &=\operatorname{clip}\big(r_0+\Delta(I,x),r_0,r_{\max}\big),\\
    K^{*}
      &=\operatorname{clip}\!\left(
        \left\lfloor r_{\mathrm{eff}}N\right\rceil,
        B,\left\lfloor r_{\max}N\right\rfloor\right).
\end{aligned}
    \label{eq:dynamic-budget}
\end{equation}
Here $B$ is a minimum-token guard and $\lfloor\cdot\rceil$ denotes nearest
integer rounding. The budget is fixed as the endpoint for the progressive
events after the first evidence read; larger uncertainty, text-like coverage,
or protected-region prevalence increases retained capacity. The implemented
response is visualized in Figure~\ref{fig:budget-risk}.

\subsection{Progressive Middle-Layer Realization}

\method{} places its pruning events in selected middle layers, allowing early
layers to contextualize the full visual sequence. Let
$\ell_1<\ell_2<\cdots<\ell_J$ be the event layers and let
$0=\alpha_0<\alpha_1<\cdots<\alpha_J=1$ be their progress coefficients.
Following the layer-wise redundancy pattern documented in
PyramidDrop~\cite{xing2025pyramiddrop}, the schedule uses a sample-specific
endpoint $K^*$ and retains more visual capacity through the earlier stages.
The temporary target interpolates from the initial count $K_0=N$ to the final
count $K^*$:
\begin{equation}
    K_j = \left\lfloor N - \alpha_j(N-K^{*}) \right\rceil.
    \label{eq:progressive-budget}
\end{equation}
The selected index sets satisfy
$\mathcal I_j\subseteq\mathcal I_{j-1}$ and $|\mathcal I_j|=K_j$, with
$K_J=K^*$. At each event, the reader is evaluated on the currently active
visual keys, the protected-first rule in Eq.~\ref{eq:protected-selection} is
applied, and the runtime compacts hidden states, attention masks, position
information, and cache metadata together. Later decoder layers and
autoregressive steps therefore operate on a shorter visual prefix while model
parameters remain fixed.

\begin{table*}[t]
\centering
\begin{tabular}{@{}llcccc@{}}
\toprule
Backbone & Method & TextVQA & OCRBench-v2 & OCRBench-mini & Keep \\
& & Soft $\uparrow$ & Official $\uparrow$ & Exact $\uparrow$ & (\%) \\
\midrule
Qwen3-VL-8B & FastV~\cite{chen2024fastv} & 0.4569 & 0.3096 & 0.5103 & 50.0 \\
& PyramidDrop~\cite{xing2025pyramiddrop} & 0.7463 & \underline{0.4168} & \underline{0.6495} & 50.0 \\
& SparseVLMs~\cite{zhang2025sparsevlm} & \underline{0.7730} & 0.3945 & 0.5619 & 50.0 \\
& IVC-Prune~\cite{sun2026ivcprune} & 0.6042 & 0.2300 & 0.4485 & 51.1 \\
& ZOO-Prune~\cite{kim2025zooprune} & 0.7190 & 0.3314 & 0.5773 & 50.0 \\
& \textbf{\method} & \textbf{0.7830} & \textbf{0.4348} & \textbf{0.6598} & 50.3 \\
\midrule
InternVL3.5-8B & FastV~\cite{chen2024fastv} & 0.3950 & 0.2325 & 0.3196 & 50.0 \\
& PyramidDrop~\cite{xing2025pyramiddrop} & \underline{0.7645} & \underline{0.3874} & \textbf{0.7113} & 50.0 \\
& SparseVLMs~\cite{zhang2025sparsevlm} & 0.5369 & 0.2527 & \underline{0.3918} & 50.0 \\
& IVC-Prune~\cite{sun2026ivcprune} & 0.1106 & 0.1201 & 0.0670 & 51.4 \\
& ZOO-Prune~\cite{kim2025zooprune} & 0.3029 & 0.1940 & 0.1753 & 50.0 \\
& \textbf{\method} & \textbf{0.7695} & \textbf{0.3942} & \textbf{0.7113} & 49.1 \\
\midrule
LLaVA-1.5-7B & FastV~\cite{chen2024fastv} & 0.3290 & 0.0828 & 0.1753 & 55.3 \\
& PyramidDrop~\cite{xing2025pyramiddrop} & 0.3442 & 0.0827 & 0.1856 & 55.3 \\
& SparseVLMs~\cite{zhang2025sparsevlm} & 0.3178 & 0.0820 & 0.1959 & 50.0 \\
& IVC-Prune~\cite{sun2026ivcprune} & 0.2248 & 0.0617 & 0.1082 & 51.8 \\
& ZOO-Prune~\cite{kim2025zooprune} & \underline{0.4408} & \underline{0.1039} & \underline{0.2371} & 50.0 \\
& \textbf{\method} & \textbf{0.4522} & \textbf{0.1040} & \textbf{0.2423} & 45.9 \\
\bottomrule
\end{tabular}
\caption{Quality comparison among pruned methods. ``Keep'' denotes average
visual-token retention on OCRBench-v2. Bold and underline mark the best and
second-best distinct pruned results; ties are bolded.}
\label{tab:main-results}
\end{table*}

\subsection{Complexity and Deployment}

Let $T$ be the number of text tokens, $L_s$ the number of decoder layers in
stage $s$, and $d$ the hidden width. If $C(n)=O(nd^2+n^2d)$ denotes the
per-layer transformer cost for sequence length $n$, the approximate prefill
cost of \method{} is
\begin{equation}
    C_{\mathrm{ET}}
    \approx \sum_{s=0}^{J}L_s C(T+K_s)
    + \sum_{j=1}^{J}O\!\left(|Q|K_{j-1}d\right).
    \label{eq:method-complexity}
\end{equation}
The first term captures the progressively shortened decoder sequence and the
second term is the partial-QK reader overhead. The scorer materializes only the
question-to-visual block, while the main forward pass stays on the model's
efficient attention path. The corresponding KV-cache footprint scales as
\begin{equation}
    M_{\mathrm{KV}}=O\!\left(d\sum_{s=0}^{J}L_s(T+K_s)\right),
    \label{eq:kv-complexity}
\end{equation}
which accounts for the stage-specific visual prefix retained by each decoder
layer. All reported FLOPs include the reader term in Eq.~\ref{eq:method-complexity}.


\section{Experiments}

\subsection{Experimental Setup}
We evaluate the decoder-side \method{} implementation on Qwen3-VL-8B and
InternVL3.5-8B~\cite{bai2025qwen3vl,wang2025internvl35}. The core benchmarks
are TextVQA, OCRBench-v2, and a reproducible 194-example OCRBench-mini
diagnostic~\cite{singh2019textvqa,liu2024ocrbench,fu2025ocrbenchv2}; the main
table reports their standard soft score, official bilingual aggregate, and
exact match, respectively. We compare with FastV, PyramidDrop, SparseVLMs,
IVC-Prune, and ZOO-Prune at comparable average retention
near 50\%~\cite{chen2024fastv,xing2025pyramiddrop,zhang2025sparsevlm,
sun2026ivcprune,kim2025zooprune}. All methods share the model, image
preprocessing, greedy decoding, and evaluation pipeline, and run with batch
size one on an NVIDIA RTX A6000. Each score is one deterministic full-split
pass, so we report observed point estimates without statistical-significance
claims. The core \method{} configurations are selected on a fixed 500-example
TextVQA training subset disjoint from the three core evaluation sets and frozen
before evaluation; no OCRBench labels are used for selection. Complete decoding
settings, hyperparameters, calibration protocol, software versions, and
per-dataset operating points are provided in the supplementary material.

\subsection{Quality at Comparable Retention}
Table~\ref{tab:main-results} compares the observed point estimates. On
Qwen3-VL-8B, \method{} improves over the strongest competing pruned row by
1.00, 1.80, and 1.03 percentage points on TextVQA, OCRBench-v2, and
OCRBench-mini, respectively, while retaining 50.3\% of visual tokens on
OCRBench-v2. On InternVL3.5-8B, it gains 0.50 and 0.68 points on TextVQA and
OCRBench-v2 while retaining 49.1\% on OCRBench-v2; it also ties the best
OCRBench-mini score. Thus, \method{} leads or ties all six core
backbone--benchmark comparisons. The
LLaVA-1.5-7B rows are a compatibility diagnostic at the reported operating
points: its \method{} variant uses a pre-LLM selector and is not part of the
two-backbone decoder-side claim~\cite{liu2024llava15}.

\begin{figure}[t]
    \centering
    \includegraphics[width=\columnwidth]{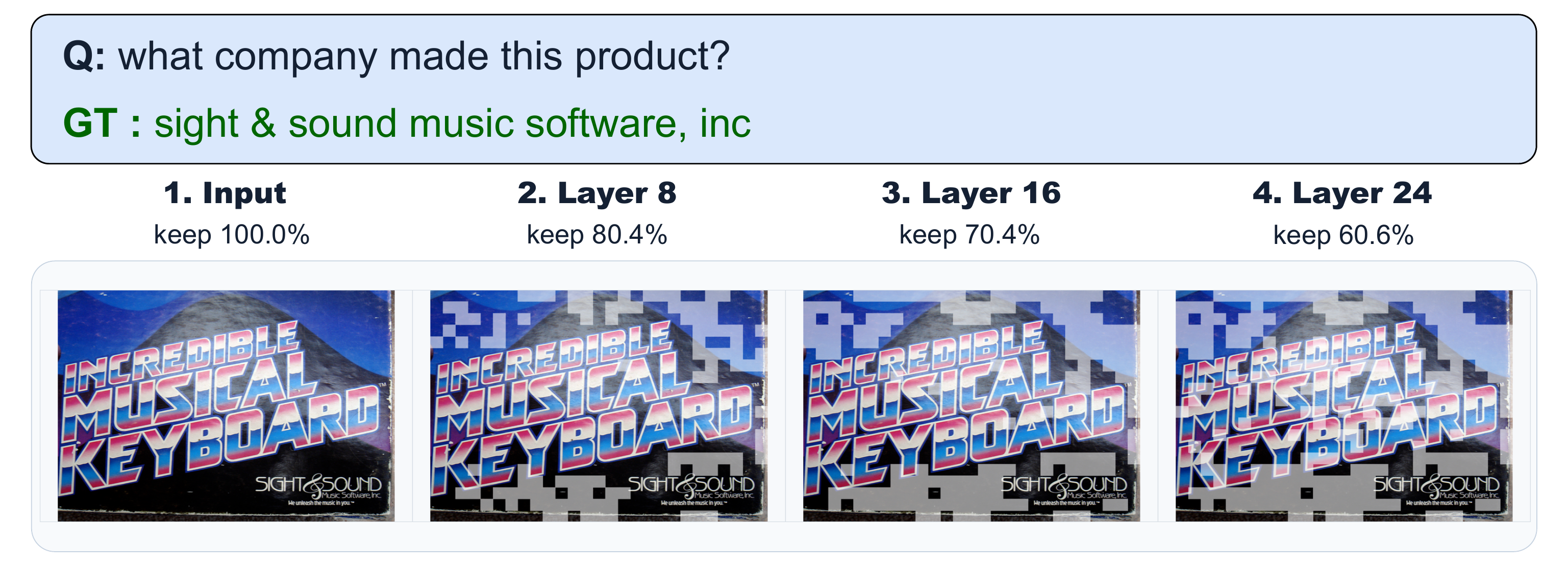}
    \par
    \includegraphics[width=\columnwidth]{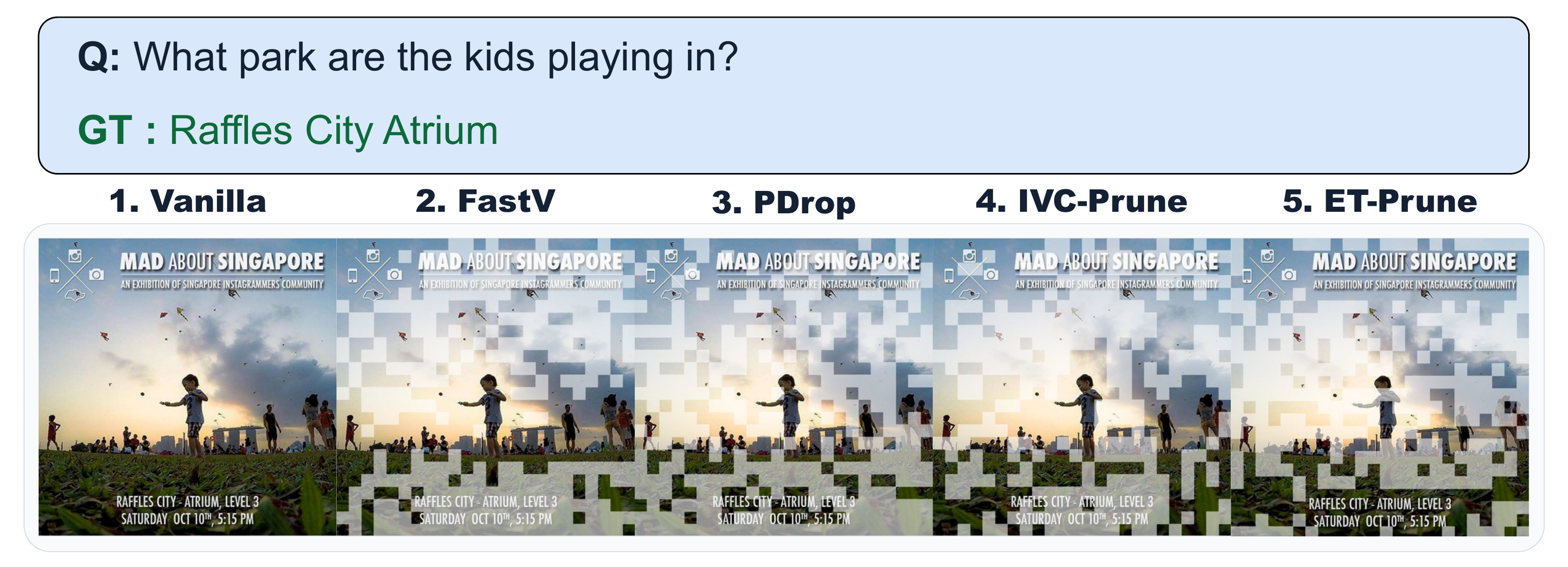}
    \par
    \caption{Qualitative TextVQA comparison. Top: \method{} progressively
    compacts the visual sequence while preserving the question-relevant product
    label. Bottom: Vanilla and \method{} answer correctly; FastV, PyramidDrop,
    and IVC-Prune lose the product-location evidence.}
    \label{fig:qualitative-comparisons}
\end{figure}

\begin{table}[t]
\centering
\setlength{\tabcolsep}{3pt}
\begin{tabular}{@{}l|ccccc@{}}
\toprule
\multirow{2}{*}{Method} & \multicolumn{5}{c}{Protocol-specific score $\uparrow$} \\
\cmidrule{2-6}
 & MMVP & Robot & RealW. & MME & MMB. \\
\midrule
Vanilla & 0.6600 & 0.5786 & 0.7033 & 2400.84 & 0.8437 \\
FastV & 0.5600 & 0.5762 & 0.6837 & 2268.52 & 0.8243 \\
PyramidDrop & 0.5800 & \underline{0.5765} & 0.6993 & 2322.49 & 0.8251 \\
SparseVLMs & 0.5467 & 0.5755 & 0.6889 & \underline{2358.22} & 0.8320 \\
IVC-Prune & \textbf{0.6200} & 0.5703 & \textbf{0.7059} & 2246.08 & \underline{0.8421} \\
ZOO-Prune & 0.5600 & 0.5469 & 0.6680 & 2261.13 & 0.8189 \\
\textbf{\method} & \underline{0.5867} & \textbf{0.5771} & \underline{0.7007} & \textbf{2371.97} & \textbf{0.8467} \\
\bottomrule
\end{tabular}
\caption{Transfer scores on Qwen3-VL-8B at nominal 50\% visual-token retention.
Robot, RealW., and MMB. denote the converted RoboVQA split, RealWorldQA, and
MMBench v1.1. Bold and underline mark the best and second-best pruned results.}
\label{tab:general-transfer}
\end{table}

\subsection{Transferability beyond OCR}
Table~\ref{tab:general-transfer} evaluates MMVP, a converted RoboVQA split,
RealWorldQA, MME, and MMBench v1.1 under their dataset-specific
protocols~\cite{tong2024mmvp,sermanet2023robovqa,xai_grok1_5,
fu2023mme,liu2024mmbench}. Among the pruned rows, \method{} ranks first on the
robot-VQA split, MME, and MMBench and second on RealWorldQA and MMVP. In
particular, its 0.5867 MMVP score remains below IVC-Prune's 0.6200. These
results characterize a quality--compression trade-off beyond OCR rather than a
universal accuracy gain. Full scoring, conversion, and retention details are
provided in the supplementary material.

\subsection{Qualitative Evidence Preservation}
Figure~\ref{fig:qualitative-comparisons} shows a representative case
from TextVQA. In the top line, we observe that the company name is preserved across layers for pruning. In the bottom line, the progressive compaction preserves the answer-related park name that
is lost by the baselines like FastV and IVC-Prune. These all suggest the strong capabilities of ET-Prune on preserving evidences.

\subsection{Compatibility efficiency}

The LLaVA-1.5-7B compatibility implementation reduces relative prefill compute
to 53.38\%, uses 1.89~T prefill FLOPs and 194.12~MB of KV-cache memory, and
reaches 0.0704~s prefill time, with second-best decode latency. Its pre-LLM
selector bypasses the decoder-side reader, so this result demonstrates
compatibility rather than the runtime of the two core implementations. The
complete efficiency table and measurement details are provided in the
supplementary material.

\begin{table}[tb]
\centering
\setlength{\tabcolsep}{4pt}
\begin{tabular*}{\columnwidth}{@{\extracolsep{\fill}}lrrr@{}}
\toprule
Target & Keep & Score $\uparrow$ & FLOPs $\downarrow$ \\
\midrule
\multicolumn{4}{@{}l}{\textit{TextVQA (soft score)}} \\
\midrule
R60 & 59.09 & 0.7917 & 70.47 \\
\textbf{R50}$^{*}$ & 49.09 & 0.7830 & 63.10 \\
R40 & 39.08 & 0.7671 & 55.75 \\
\addlinespace
\midrule
\multicolumn{4}{@{}l}{\textit{OCRBench-v2 (official aggregate)}} \\
\midrule
R60 & 61.29 & 0.4480 & 76.00 \\
\textbf{R50}$^{*}$ & 50.28 & 0.4348 & 68.98 \\
R40 & 39.91 & 0.4188 & 62.39 \\
\bottomrule
\end{tabular*}
\caption{Retention-target ablation on Qwen3-VL-8B. Keep and FLOPs denote
dataset-level mean visual-token retention and relative prefill compute (\%),
respectively; FLOPs includes the reader overhead. R50$^{*}$ is the frozen main
operating point.}
\label{tab:retention-ablation}
\end{table}

\subsection{Retention-target Ablation}
Table~\ref{tab:retention-ablation} compares three adjacent budget targets. From
R60 to R50, relative prefill compute decreases by 7.37 points on TextVQA and
7.02 points on OCRBench-v2, while the corresponding scores decrease by 0.87
and 1.32 points, respectively. R40 reduces compute further but incurs larger
quality losses, placing the frozen R50 configuration at an intermediate
quality--compute operating point. Additional budget targets, layer-schedule
ablations, and metric breakdowns are provided in the supplementary material.


\section{Conclusion}

We introduced \method, a training-free visual-token pruner for MLLMs on text-rich tasks. It allocates capacity with question-to-visual partial attention, a
spatial safeguard, and a risk-calibrated floor. Across extensive backbones and benchmarks, its observed point estimates preserve more quality than
representative pruning baselines at comparable average token counts.
Cross-benchmark transfer and supplementary Qwen3-VL scaling results further
support evidence-aware allocation beyond a single model scale and OCR-only
tasks.

\paragraph{Limitations.}
The current batch-one implementation requires model-specific bookkeeping for
compacted position and cache states. Its lightweight spatial prior is not a text detector,
and budget parameters may need retuning for new image resolutions or tokenizers.
Future work should study batched dynamic compaction and broader document and
video calibration.

\section{Declaration of LLM Usage}
In this paper, we use LLMs (mainly GPT-5.6) to polish our language and find typos, mistakes and so on. We commit to using LLMs in compliance with all applicable AAAI 2027 policies and will not engage in any prohibited practices, such as using LLMs to generate hallucinated citations.


\bibliography{aaai2027}
\end{document}